%% file: main.tex
\documentclass[10pt,twocolumn,letterpaper]{article}
\usepackage[pagenumbers]{cvpr}
\usepackage{placeins}
\usepackage{graphicx,amsmath,amssymb,bm,booktabs,multirow,microtype}
\usepackage[ruled,vlined,linesnumbered]{algorithm2e}
\SetAlFnt{\footnotesize}
\definecolor{cvprblue}{rgb}{0.21,0.49,0.74}
\usepackage[breaklinks,colorlinks,allcolors=cvprblue]{hyperref}
\hypersetup{
  pdftitle={Learning Goal-Conditioned Subgoals through Frozen Hierarchical World Models},
  pdfauthor={Kaibo Xiang and Qingfeng Liu},
  pdfsubject={Goal-conditioned planning through frozen hierarchical world models}
}
\graphicspath{{fig/}}
\newcommand{\dist}{\operatorname{MAE}}
\newcommand{\suppref}[1]{Supp.~Sec.~#1}
\newcommand{\supprefs}[1]{Supp.~Secs.~#1}
\AtBeginDocument{\crefname{algorithm}{Alg.}{Algs.}\Crefname{algorithm}{Algorithm}{Algorithms}}

\title{Learning Goal-Conditioned Subgoals through\\Frozen Hierarchical World Models}
\author{Kaibo Xiang\qquad Qingfeng Liu\thanks{Corresponding author.}\\[4pt]
  \normalsize Department of Industrial and Systems Engineering, Hosei University\\[2pt]
  \normalsize \href{mailto:kaibo.xiang.4h@stu.hosei.ac.jp}{\texttt{kaibo.xiang.4h@stu.hosei.ac.jp}}\qquad
  \href{mailto:qliu@hosei.ac.jp}{\texttt{qliu@hosei.ac.jp}}}
\date{}

\newcommand{\main}[1]{#1 of the main paper}

\begin{document}
\maketitle
\input{paper_body}

\FloatBarrier
\section*{Acknowledgments}
This work was supported by JSPS KAKENHI (Grant Number JP25K05040).

\section*{LLM Usage Statement}
\input{AI_ASSISTANCE_DISCLOSURE}

{\small
\bibliographystyle{ieeenat_fullname}
\bibliography{references}
}

\clearpage
\appendix
\input{supplement}
\end{document}

%% file: paper_body.tex
\begin{abstract}
Hierarchical Planning with Latent World Models (HWM) uses high-level search to generate subgoals for low-level action planning. We augment this pretrained hierarchy with a lightweight goal-conditioned head to guide action selection while reducing the computational cost of repeated high-level search. Given the current latent state and goal, the head produces a conditioning vector; the frozen high-level predictor uses this vector to generate a latent subgoal that guides the retained low-level planner in selecting actions. We train only the head through the frozen predictor on recorded trajectories, encouraging predicted intermediate states to match recorded states and the predicted endpoint to approach the goal. On four fixed simulated visual-control benchmarks, our planner improves the mean task success rate over HWM search by 19--36 percentage points on three benchmarks and matches it on the fourth. It also achieves higher mean success than two-level search with the Hierarchical Joint-Embedding Predictive Architecture (H-JEPA) on all four. With the released baseline search settings, batched timing runs show a reduction of more than two orders of magnitude in high-level solver time and \mbox{$1.28$--$4.39\times$} end-to-end speedups over HWM search.
\end{abstract}

\section{Introduction}
\label{sec:intro}

Visual world models support control by predicting the consequences of actions in a compact latent space~\cite{ha2018world,hafner2019planet,hansen2022tdmpc,zhou2024dinowm,maes2026lewm}. Long-horizon planning remains difficult because the action search grows with temporal depth and prediction errors accumulate. Hierarchical Planning with Latent World Models (HWM)~\cite{zhang2026hwm} decomposes control across temporal scales within a shared latent space: its high-level planner predicts latent trajectories and passes the first predicted state to a low-level model predictive control (MPC) planner. We call the predicted high-level states \emph{waypoints}; the first, handed to the low level, is the \emph{subgoal}.

In the HWM configuration distributed with the Hierarchical Joint-Embedding Predictive Architecture (H-JEPA) release~\cite{zhang2026hjepa} and used throughout this paper, the high-level planner optimizes candidate macro-action sequences by repeatedly rolling out and differentiating through the coarse (high-level) predictor (the original HWM paper uses sampling-based solvers; \suppref{D}). Depending on the environment, the released configurations allow 9{,}000--72{,}000 candidate-trajectory updates per decision (\cref{tab:budget}); even with early stopping, high-level search takes 18--71\% of end-to-end evaluation time in our timing runs (\cref{sec:cost}), and it is repeated whenever the high level is active.

\begin{figure*}[t]
  \centering
  \includegraphics[width=0.9\textwidth]{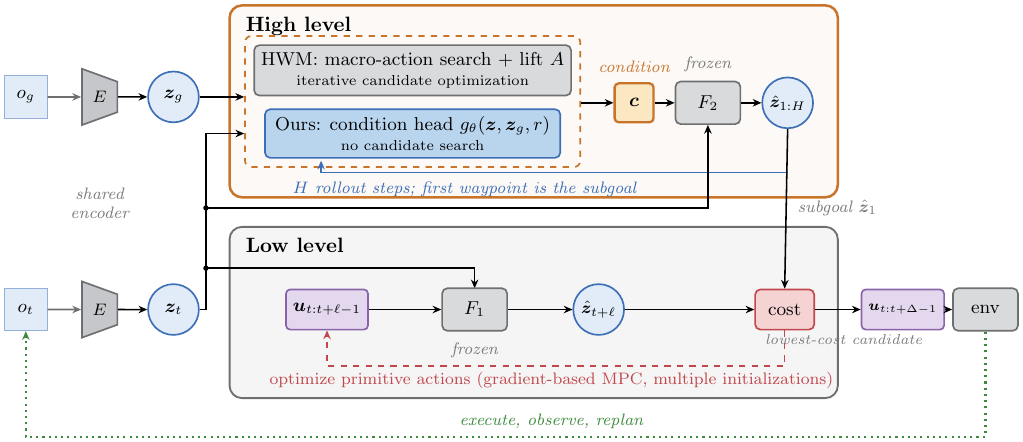}
  \caption{\textbf{Learning conditions for HWM's frozen hierarchy.} Encoder $E$ maps the current and goal observations to latent states $\bm z_t$ and $\bm z_g$. HWM searches over macro-actions and applies the lift $A$ to obtain conditions for the coarse predictor $F_2$; our head $g_\theta$ (blue) supplies a condition $\bm c$ from the state, goal, and remaining-step input $r$. Of the $H$ predicted waypoints, the first, $\hat{\bm z}_1$, becomes the low-level subgoal. The retained low-level MPC planner optimizes $\ell=5L$ primitive actions $\bm u_{t:t+\ell-1}$ through the fine predictor $F_1$ by gradient-based search (dashed red loop; $L$ is the number of five-action blocks, \cref{sec:inference}), then executes a prefix of length $\Delta$. Only $g_\theta$ is updated during head training.}
  \label{fig:overview}
\end{figure*}

We build on HWM's hierarchy by learning to supply its coarse predictor with goal-dependent conditions. A small \emph{condition head} $g_\theta$, with trainable parameters $\theta$, maps a latent state, the goal, and a remaining-step input to a condition vector for the frozen coarse predictor. The predictor then generates the next waypoint, while the low-level planner continues to optimize primitive actions through the frozen fine predictor (\cref{fig:overview}). The first head--predictor evaluation yields the subgoal, without candidate macro-action optimization.

Recorded trajectories provide both hindsight goals and evidence of state changes realized by executed actions. We use this evidence to train the head through the frozen predictor (\cref{sec:train}): a teacher-forced loss fits observed local transitions, a rollout loss keeps recursively predicted states close to the recorded trajectory, and an endpoint loss gives extra weight to goal proximity. The intermediate losses thus constrain the predicted trajectory with this evidence, while the low-level planner searches for actions toward the resulting subgoal. All targets are encoded observations; no macro-action labels or search-generated plans are required.

We evaluate on four simulated visual-control benchmarks using the HWM checkpoints distributed with the H-JEPA release, in which both levels share one latent space. HWM search therefore uses exactly the frozen backbone our head is attached to, whereas two-level H-JEPA search, which learns a separate representation at each level, is a separately trained hierarchical baseline; \cref{sec:experiments} compares our planner with both, with the single-level LeWM planner, and with linear latent waypoints.

Our contributions are as follows:
\begin{itemize}\setlength\itemsep{1pt}
 \item We propose a subgoal-generation method that uses a pretrained hierarchical world model to guide action planning. A learned condition head supplies goal-dependent conditions to the frozen high-level predictor, eliminating repeated high-level candidate macro-action optimization at deployment while retaining low-level action planning.
 \item Our method trains the condition head using intermediate states and endpoints from recorded trajectories as supervision, encouraging predicted intermediate states to match recorded states and the predicted endpoint to approach the goal. Training requires neither search-generated plans nor macro-action labels.
 \item Our method demonstrates strong control performance and computational efficiency on four fixed benchmark sets: compared with HWM search, it raises mean success by 19--36 percentage points on three sets and matches it on the fourth, with \mbox{$1.28$--$4.39\times$} end-to-end speedups in batched timing runs.
\end{itemize}

\section{Related Work}

\paragraph{Latent world models and hierarchical planning.}
World Models~\cite{ha2018world}, PlaNet~\cite{hafner2019planet}, Dreamer~\cite{hafner2020dreamer,hafner2021dreamerv2,hafner2023dreamerv3}, and temporal-difference MPC (TD-MPC)~\cite{hansen2022tdmpc,hansen2024tdmpc2} demonstrate predictive control and policy learning in latent spaces. DINO-WM~\cite{zhou2024dinowm} and LeWorldModel (LeWM)~\cite{maes2026lewm} plan in learned feature spaces without pixel reconstruction. For navigation, Bar et al.~\cite{bar2025nwm} plan with action-conditioned video predictions. The joint-embedding predictive architecture (JEPA)~\cite{lecun2022path} motivates prediction in representation space; image JEPA (I-JEPA)~\cite{assran2023ijepa} and video JEPA (V-JEPA)~\cite{bardes2024vjepa,assran2025vjepa2} instantiate it for images and video. HWM~\cite{zhang2026hwm} and H-JEPA~\cite{zhang2026hjepa} develop predictive hierarchies across temporal scales (\cref{sec:intro}).

\paragraph{Test-time optimization.}
Sampling-based planners use cross-entropy optimization~\cite{rubinstein2004cem,deboer2005tutorial,pinneri2020icem} or model predictive path integral control~\cite{williams2016mppi} to search over action sequences. PETS~\cite{chua2018pets} and PDDM~\cite{nagabandi2019pddm} apply such sampling-based optimization to learned dynamics models, while differentiable MPC~\cite{amos2018diffmpc} connects optimization-based control with learning. Our head reduces repeated high-level optimization while retaining the low-level planner.

\paragraph{Goal-conditioned and hierarchical policies.}
Options~\cite{sutton1999options,bacon2017optioncritic}, FeUdal Networks~\cite{vezhnevets2017feudal}, and Director~\cite{hafner2022director} use temporal abstraction in reinforcement learning; off-policy hierarchical methods learn subgoal policies with data reuse and hindsight~\cite{nachum2018hiro,levy2019hac}. Hindsight relabeling~\cite{andrychowicz2017her}, visual goal representations~\cite{nair2018rig}, goal-conditioned supervised learning~\cite{ghosh2021gcsl}, and offline goal-conditioned methods~\cite{eysenbach2022contrastive,park2023hiql} learn goal-conditioned policies from relabeled or offline data. Our head also learns from hindsight goals, but its output conditions a frozen coarse predictor rather than directly specifying a primitive action~\cite{ghosh2021gcsl} or a latent subgoal as in HIQL~\cite{park2023hiql}.

\paragraph{Amortizing planning.}
Amortized optimization learns a mapping from problem instances to approximate solutions, avoiding repeated per-instance optimization~\cite{amos2022amortized}. AlphaZero and MuZero~\cite{silver2018alphazero,schrittwieser2020muzero} distill search outputs into networks; our head instead learns from encoded observations, with no search in the training loop. Closest to our setting, Nguyen et al.~\cite{nguyen2026latent} amortize planning in the frozen single-level LeWM~\cite{maes2026lewm} latent space with a goal-conditioned inverse-dynamics model (GC-IDM), a small multilayer perceptron that maps the current latent, the goal latent, and the remaining horizon to the next action by regression onto recorded actions. Our head takes the same three inputs but differs in its output (a condition vector for a frozen coarse predictor rather than an action), its supervision (latent-state errors backpropagated through that predictor rather than action labels), and its role (the high level of a two-level planner whose low-level MPC is retained).

\section{Method}
\label{sec:method}

\subsection{Frozen two-level world model}

Let $o_t$ be the observation at environment step $t$, $o_g$ the goal observation, and $E$ the frozen observation encoder. Their latent states are $\bm z_t=E(o_t)$ and $\bm z_g=E(o_g)$ in $\mathbb R^d$, where $d$ is the latent dimension. The frozen predictors $F_1$ and $F_2$, both transformers, operate at fine and coarse temporal scales. These modules are inherited from the pretrained backbone; our training adds the condition head and updates only its parameters. Both predictors operate on the same encoder output, with no separate high-level encoder; we call this released configuration \emph{shared-latent} HWM, in contrast to H-JEPA, which learns a representation per level. A fine transition ($F_1$) spans 5 environment steps and a coarse transition ($F_2$) 10. Below, $t$ indexes environment time, whereas $j$ and $k$ index coarse positions within a trajectory segment or a newly initialized rollout.

The coarse predictor $F_2$ takes, alongside each state, a \emph{condition vector} $\bm c\in\mathbb R^d$. In HWM search, a frozen macro-action embedding (the \emph{lift}) $A:\mathbb R^4\to\mathbb R^d$ maps a four-dimensional macro-action $\bm a$ to $\bm c=A(\bm a)$. Let $H$ be the current high-level horizon (number of coarse transitions) and $\bm a_{0:H-1}=(\bm a_0,\ldots,\bm a_{H-1})$ a candidate sequence. The rollout operator $\operatorname{Roll}_{F_2}$ returns the predicted state sequence $\bar{\bm z}_{0:H}$ initialized at $\bar{\bm z}_0=\bm z_t$, with $A$ applied to each macro-action. In the HWM configuration distributed with the H-JEPA release, the high-level cost is the squared Euclidean (L2) distance between the terminal predicted state and the goal, $J(\bar{\bm z}_{0:H},\bm z_g)=\|\bar{\bm z}_H-\bm z_g\|_2^2$, so the search seeks
\begin{equation}
 \bm a^*_{0:H-1}=\operatorname*{arg\,min}_{\bm a_{0:H-1}}
 J\!\left(\operatorname{Roll}_{F_2}
 (\bm z_t,A(\bm a_{0:H-1})),\bm z_g\right).
 \label{eq:search}
\end{equation}
The released solver uses gradient updates with early stopping (\suppref{D} compares it with the original HWM solver). Candidates are initialized from a Gaussian fitted to training macro-actions and clipped to their empirical per-coordinate bounds~\cite{zhang2026hjepa}. Let $N$ be the number of candidates and $T_{\max}$ the maximum number of optimization updates. \Cref{tab:budget} lists the initial budgets, with $NT_{\max}$ bounding the number of candidate-trajectory updates.

\begin{table}[t]
\caption{\textbf{Initial high-level search budget.} $N$: candidate trajectories; $T_{\max}$: optimization-update limit; $H$: initial high-level horizon in coarse transitions (it decreases along the schedule, \suppref{D}); $NT_{\max}$ is an upper bound because of early stopping. Our planner replaces candidate optimization by one head--predictor call per coarse transition; the first call determines the subgoal (\cref{sec:inference}).}
\label{tab:budget}
\centering\small\setlength{\tabcolsep}{4pt}
\begin{tabular}{lrrrr}
\toprule
Environment & $N$ & $T_{\max}$ & $H$ & $NT_{\max}$\\
\midrule
Cube & 100 & 90 & 2 & 9{,}000\\
Push-T & 400 & 90 & 8 & 36{,}000\\
FourRoom & 800 & 90 & 12 & 72{,}000\\
AntMaze & 200 & 90 & 9 & 18{,}000\\
\bottomrule
\end{tabular}
\end{table}

\subsection{Condition head}

Let $g_\theta$ be the condition head with trainable parameters $\theta$. For a latent state $\bm z$, goal $\bm z_g$, and nonnegative integer remaining-step input $r$, it produces
\begin{equation}
 \bm c=g_\theta(\bm z,\bm z_g,r)\in\mathbb R^d.
 \label{eq:condition}
\end{equation}
The head is a multilayer perceptron. Its input is the concatenation $[\bm z;\bm z_g;\mathrm{PE}(r)]$, where $\mathrm{PE}$ is a 16-dimensional sinusoidal positional encoding (\suppref{C}). Layer normalization and two 512-unit hidden layers with Gaussian error linear unit (GELU) activations precede the output layer. The head directly produces a condition in $\mathbb R^d$, bypassing the macro-action lift $A$; the frozen $F_2$ then maps that condition and the state context to a waypoint. This parameterization keeps the head independent of the macro-action pathway: the head is trained without macro-action labels, and its output is passed to $F_2$ without the lift $A$ during both training and deployment. \Cref{sec:ablation} compares this parameterization with a four-dimensional output passed through $A$.

The head acts pointwise on states, while $F_2$ attends over an ordered context of up to three state/condition pairs. At prediction step $k$, define the index set $\mathcal I_k=\{\max(0,k-2),\ldots,k\}$ and the state sequence $C_k=(\bm z_j)_{j\in\mathcal I_k}$, ordered by increasing $j$. Let $r_{k,j}$ be the remaining-step input assigned to context position $j$ at prediction step $k$, $\bm r_k=(r_{k,j})_{j\in\mathcal I_k}$ the corresponding ordered vector, and $\bm c_{k,j}=g_\theta(\bm z_j,\bm z_g,r_{k,j})$. Overloading notation, we let $F_2(\cdot,\cdot)$ denote the predictor's output at the last context position, i.e., the next latent state, and define the head-conditioned transition operator
\begin{equation}
 P_\theta(C_k,\bm z_g,\bm r_k)=F_2\big(C_k,(\bm c_{k,j})_{j\in\mathcal I_k}\big).
 \label{eq:operator}
\end{equation}
With predicted states $\hat{\bm z}_j$, we use the same index set to form $\hat C_k=(\hat{\bm z}_j)_{j\in\mathcal I_k}$. HWM search uses its released one-state rollout context, whereas the head is trained with contexts of up to three states (\cref{sec:inference}).

\subsection{Learning from hindsight trajectories}
\label{sec:train}

\begin{figure*}[t]
 \centering
 \includegraphics[width=0.9\textwidth]{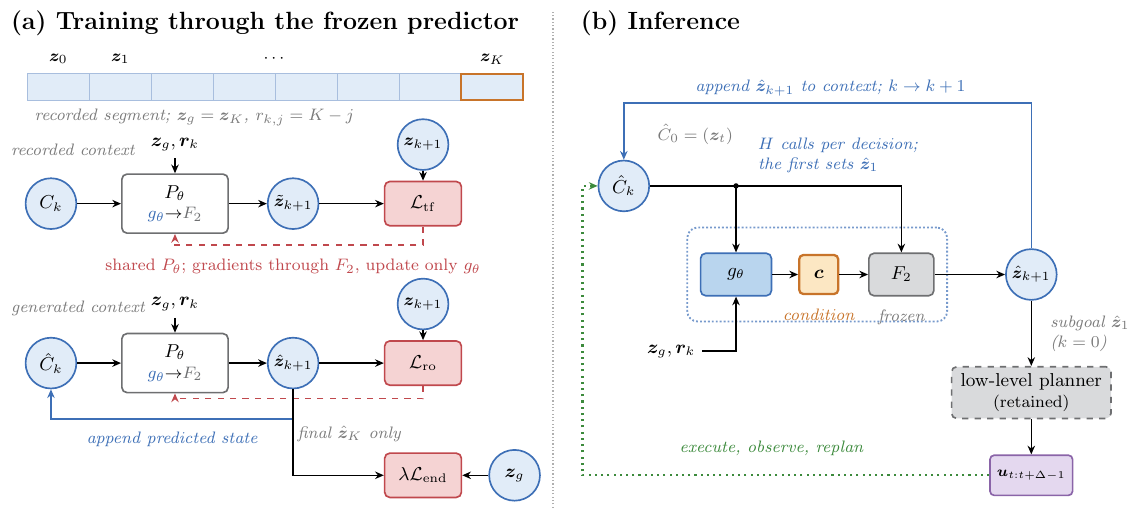}
 \caption{\textbf{Learning and using predictor conditions.} (a) The head is trained through the frozen predictor on recorded segments $\bm z_0,\ldots,\bm z_K$ with hindsight goal $\bm z_g=\bm z_K$. Both passes share $P_\theta$ ($g_\theta$ followed by $F_2$): recorded contexts $C_k$ give teacher-forced predictions $\tilde{\bm z}_{k+1}$ ($\mathcal L_{\mathrm{tf}}$), generated contexts $\hat C_k$ give autoregressive predictions $\hat{\bm z}_{k+1}$ ($\mathcal L_{\mathrm{ro}}$), and $\lambda\mathcal L_{\mathrm{end}}$ uses only $\hat{\bm z}_K$. Gradients from all three terms propagate through $F_2$ (dashed red; the endpoint term shares the rollout path), while only the parameters of $g_\theta$ are updated. (b) At inference, the goal and remaining-step vector $\bm r_k$ condition predictions from context $\hat C_k$; for $H>1$ the rollout makes $H$ head and $H$ predictor calls, and the first call yields the subgoal $\hat{\bm z}_1$ for the retained low-level MPC planner.}
 \label{fig:method}
\end{figure*}

Let $K$ be the number of coarse transitions in a training segment. We cache embeddings of the recorded trajectories and sample segments $(\bm z_0,\ldots,\bm z_K)$ of $K+1$ states, with the final state relabeled as the hindsight goal $\bm z_g=\bm z_K$. Training uses remaining-step inputs $r_{k,j}=K-j$, so the input attached to the most recent state counts down \mbox{$K,K-1,\ldots,1$}.

For latent vectors $\bm x,\bm y\in\mathbb R^d$, we measure prediction error with the mean absolute error over latent coordinates, $\dist(\bm x,\bm y)=\frac1d\sum_{i=1}^{d}|x_i-y_i|$. All training losses use MAE, whereas the search cost in \cref{eq:search} uses squared L2 distance. The following equations describe one segment; training averages the objective over a mini-batch.

\paragraph{Teacher-forced loss.}
The teacher-forced loss predicts each transition from its recorded context $C_k$ (upper row of \cref{fig:method}a), giving $\tilde{\bm z}_{k+1}=P_\theta(C_k,\bm z_g,\bm r_k)$:
\begin{equation}
 \mathcal L_{\mathrm{tf}}=\frac1K\sum_{k=0}^{K-1}
 \dist\!\left(\tilde{\bm z}_{k+1},\bm z_{k+1}\right).
 \label{eq:tf}
\end{equation}
It encourages the head to produce conditions under which the frozen predictor's next-state prediction matches the recorded next state. The inputs here are recorded states, so each transition is supervised without errors from earlier predictions.

\paragraph{Rollout loss.}
To train on self-generated contexts, we set $\hat{\bm z}_0=\bm z_0$ and predict recursively
\begin{equation}
 \hat{\bm z}_{k+1}=P_\theta(\hat C_k,\bm z_g,\bm r_k),
 \quad k=0,\ldots,K-1.
 \label{eq:recurrence}
\end{equation}
The rollout loss $\mathcal L_{\mathrm{ro}}$ compares every generated future state with its recorded counterpart:
\begin{equation}
 \mathcal L_{\mathrm{ro}}=\frac1K\sum_{k=1}^{K}
 \dist(\hat{\bm z}_k,\bm z_k).
 \label{eq:ro}
\end{equation}
After the first transition, the head and predictor also receive previously predicted states in their context. $\mathcal L_{\mathrm{ro}}$ therefore penalizes the drift that accumulates along the generated trajectory. Gradients propagate through the complete rollout without detaching intermediate predictions.

\paragraph{Endpoint loss.}
The endpoint loss $\mathcal L_{\mathrm{end}}$ additionally penalizes the distance between the final prediction $\hat{\bm z}_K$ and the hindsight goal:
\begin{equation}
 \mathcal L_{\mathrm{end}}=\dist(\hat{\bm z}_{K},\bm z_g).
 \label{eq:end}
\end{equation}

\paragraph{Total training objective.}
Let $\lambda\geq0$ weight the endpoint loss. Our default objective is
\begin{equation}
 \mathcal L=\mathcal L_{\mathrm{tf}}+\mathcal L_{\mathrm{ro}}
             +\lambda\mathcal L_{\mathrm{end}},\qquad\lambda=1.
 \label{eq:objective}
\end{equation}
Since $\hat{\bm z}_K$ also appears in $\mathcal L_{\mathrm{ro}}$, its total weight is $1/K+\lambda$. Endpoint supervision alone supplies no direct targets for intermediate states. Recorded intermediate states show transitions realized during execution; supervising them encourages predicted trajectories to follow those observed changes, while the extra endpoint term emphasizes proximity to the goal. At deployment, the frozen high-level predictor generates the subgoal that guides the retained low-level action search.

All three terms update only $\theta$, with gradients passing through the frozen $F_2$ and, for the autoregressive terms, through earlier predictions. Head training uses encoded observation sequences (with proprioception for AntMaze), without action labels or search-generated plans. We optimize latent prediction errors and evaluate task success as the downstream outcome of the complete planner. \Cref{sec:ablation} and \suppref{E} examine the individual loss terms on Cube and Push-T; \cref{alg:train} summarizes training.

We fit the head with AdamW~\cite{loshchilov2019adamw} and select checkpoints by autoregressive endpoint MAE on fixed validation windows of $K_{\mathrm{val}}$ transitions, where $K_{\mathrm{val}}$ is the environment's default $K$ even when the training length is ablated (\cref{sec:protocol}).

\begin{algorithm}[t]
\caption{Learning a condition head through a frozen predictor}\label{alg:train}
\KwIn{cached latent trajectories; fixed validation windows of $K_{\mathrm{val}}$ transitions; frozen $F_2$; training length $K$; endpoint weight $\lambda$; number of updates $S=20{,}000$}
Initialize $\theta$, AdamW, and cosine learning-rate schedule; set best score $e^*=\infty$ and saved parameters $\theta^*=\theta$\;
\For{$s=1,\ldots,S$}{
 Sample a mini-batch of $K$-transition segments\;
 Set each goal to $\bm z_K$ and remaining-step inputs to $r_{k,j}=K-j$\;
 Predict all $K$ teacher-forced transitions from recorded contexts $C_k$ using \cref{eq:operator}\;
 Initialize $\hat{\bm z}_0=\bm z_0$\;
 \For{$k=0,\ldots,K-1$}{
  Form $\hat C_k$ from indices $\mathcal I_k$ and predict $\hat{\bm z}_{k+1}$ using \cref{eq:recurrence}\;
 }
 Average \cref{eq:objective} over the mini-batch\;
 Backpropagate through $F_2$ and the full rollout; clip the head gradient; update only $\theta$ and advance the learning-rate schedule\;
 Every $2{,}000$ updates: compute validation endpoint MAE $e$ over $K_{\mathrm{val}}$ transitions (inputs $K_{\mathrm{val}}-j$); if $e<e^*$, save $\theta^*\leftarrow\theta$, $e^*\leftarrow e$\;
}
\KwOut{validation-selected condition head $g_{\theta^*}$}
\end{algorithm}

\subsection{Inference and hierarchical control}
\label{sec:inference}

\Cref{alg:plan} and \cref{fig:method}b describe deployment with $\theta=\theta^*$. Let $B$ be the environment-step budget and $t$ the number of steps executed so far; schedules $H(t)$ and $L(t)$ specify the high-level horizon and number of five-action blocks planned by the low level (\suppref{D}). Each replanning step encodes the current and goal observations. For $H=H(t)>1$, we set $\hat{\bm z}_0=\bm z_t$ and recursively apply $P_\theta$ for $H$ transitions. Remaining-step inputs are $r_{k,j}=H-j$, except that the implementation sets all inputs to zero on the terminal call, so the latest-state inputs are $H,\ldots,2,0$ rather than ending at one as in training (\suppref{D}); this terminal convention does not affect $\hat{\bm z}_1$.

Only the first waypoint becomes the low-level subgoal: $\hat{\bm z}_1=F_2(\bm z_t,g_\theta(\bm z_t,\bm z_g,H))$, computed with a single-state context and the head's goal and horizon inputs. For $H>1$, the implementation still performs $H$ head and $H$ predictor calls (each head call evaluates $g_\theta$ on every retained context position, and the terminal call recomputes them with zero remaining-step inputs); reported timings include all of them. For $H=1$, the high level is bypassed and $\bm z_g$ is used directly. With $L=L(t)$, the retained low-level planner optimizes $5L$ primitive actions through $F_1$ to minimize squared latent distance to the target, executes a prefix of $\Delta=\min(10,5L,B-t)$ actions, and replans from the new observation; \suppref{D} specifies the low-level cost windows, solver settings, and schedules. At inference, differentiation is confined to low-level action search through $F_1$.

\begin{algorithm}[t]
\caption{Hierarchical control with learned conditions}\label{alg:plan}
\KwIn{frozen $E,F_1,F_2$; selected $g_{\theta^*}$; goal $o_g$; step budget $B$; schedules $H(t),L(t)$}
Set executed-step count $t=0$\;
\While{$t<B$}{
 Observe $o_t$; encode $\bm z_t=E(o_t)$ and $\bm z_g=E(o_g)$; set $H=H(t)$ and $L=L(t)$\;
 \eIf{$H>1$}{
 Initialize $\hat{\bm z}_0=\bm z_t$\;
 \For{$k=0,\ldots,H-1$}{
 Form $\hat C_k=(\hat{\bm z}_j)_{j\in\mathcal I_k}$ in temporal order\;
 Set $r_{k,j}=H-j$ for $j\in\mathcal I_k$; if $k=H-1$, set all remaining-step inputs to zero\;
 Predict $\hat{\bm z}_{k+1}=P_\theta(\hat C_k,\bm z_g,\bm r_k)$\;
 }
 Use $\hat{\bm z}_1$ as the low-level subgoal\;
 }{Use $\bm z_g$ as the low-level subgoal\;}
 Optimize $5L$ primitive actions through $F_1$ toward the subgoal, using the low-level cost and settings in \suppref{D}\;
 Set $\Delta=\min(10,5L,B-t)$; execute the first $\Delta$ actions and set $t\leftarrow t+\Delta$\;
}
\end{algorithm}

\section{Experiments}
\label{sec:experiments}

\subsection{Evaluation protocol}
\label{sec:protocol}

\paragraph{Tasks and backbones.}
We use the four fixed 50-task sets distributed with H-JEPA: Cube from OGBench~\cite{park2025ogbench}, Push-T~\cite{chi2023diffusionpolicy}, and FourRoom with distractors and Visual AntMaze~\cite{zhang2026hjepa} (hereafter FourRoom and AntMaze). Episode budgets are 30, 90, 210, and 350 environment steps, respectively, and the latent dimension is $d=192$ for Cube, Push-T, and FourRoom and $d=384$ for AntMaze. We use three released HWM backbones (seeds 42, 43, and 44) with three head initializations (0, 1, and 2) for each backbone. For AntMaze, head training and control apply the same proprioceptive normalization. Head training uses the released training trajectories. Checkpoint selection uses the public validation trajectories for Cube, Push-T, and AntMaze, and 150 separately collected validation trajectories for FourRoom. \suppref{B} specifies the data sources and their use in selection and task construction. \suppref{J} reports additional Cube and Push-T evaluations using backbones trained from scratch and disjoint source-trajectory groups for training, head-checkpoint selection, and testing.

\paragraph{Training and selection.}
All environments share the head architecture and default hyperparameters, fixed before inspecting results: AdamW, learning rate $10^{-3}$, cosine decay, 20{,}000 updates, batch size 256, and gradient clipping at 1.0 (\suppref{C}). The training segment length $K$ is 2, 8, 12, and 9 for Cube, Push-T, FourRoom, and AntMaze, equal to the initial high-level horizon in \cref{tab:budget}; we assess sensitivity to $K$ in \cref{sec:ablation}. We select the checkpoint with the lowest autoregressive endpoint MAE on 256 fixed validation windows, evaluated every 2{,}000 updates. Task success is evaluated after selection and is not a selection criterion.

\paragraph{Baselines.}
HWM search uses the same frozen encoder and predictors as our method. Two-level H-JEPA search and LeWM~\cite{maes2026lewm}, a single-level planner, use their separately trained checkpoints at the same backbone seeds. The \emph{linear-waypoints} baseline replaces the high level with equally spaced latent waypoints between the current state and the goal and uses the same low-level planner.

\paragraph{Comparison scope.}
Within each benchmark, all methods use the same tasks and budgets and evaluate one task at a time (chunk size one, i.e., unbatched solver calls). The released low-level planner includes an auxiliary loss matching its action embedding to the high-level macro-action, with weight $\beta=0.2$ on Cube and Push-T and $\beta=0$ on FourRoom and AntMaze. Because our head produces no macro-action, success experiments use $\beta=0$ for all hierarchical methods, a deviation from the released Cube and Push-T setting; the timing runs in \cref{sec:cost} keep $\beta=0.2$ for HWM search on Cube and Push-T. HWM search uses its released one-state rollout context and the budgets in \cref{tab:budget}; all other solver settings are the released defaults. Two-level H-JEPA search uses the same high-level budgets and solver settings as HWM search, and LeWM uses its released single-level planner configuration (\suppref{D}). Task identities are fixed, but the released stochastic reset does not guarantee identical realized initial states across runs (the two panels of \cref{fig:qual_ant} differ slightly). Evaluation consumes the full budget and records whether success is reported at any executed step.

\paragraph{Replication and uncertainty.}
For our planner, we average three head runs within each backbone and report the mean and standard error (SE) across the three backbone means; for baselines (one run per backbone), the SE is computed across the three runs. On the fixed task set, each SE is thus computed across three backbone-level means and does not estimate task-sampling uncertainty. Evaluation retains the released stochastic reset behavior.

\begin{table*}[t]
\caption{\textbf{Task success (\%) on the fixed 50-task benchmark sets.} Mean $\pm$ SE across three backbone seeds; for ours, the three head initializations are averaged within each backbone first. All hierarchical methods use matching weight $\beta=0$ and chunk size one (\cref{sec:protocol}). H-JEPA search uses the two-level configuration; LeWM is a single-level planner. Best mean per row in bold; on a tie both are bold.}
\label{tab:main}
\centering\small\setlength{\tabcolsep}{6pt}
\input{tables/main_results.tex}
\end{table*}

\subsection{Task success}

\Cref{tab:main} reports task success. Mean success improves over HWM search from 51.3\% to 70.4\% on Cube, from 72.0\% to 100.0\% on FourRoom, and from 35.3\% to 71.1\% on AntMaze; on Push-T, both reach 48.7\%. Per backbone, our planner exceeds HWM search on all three backbones for Cube ($+13.3$ to $+23.3$ percentage points), FourRoom ($+20.0$ to $+42.0$), and AntMaze ($+32.0$ to $+40.7$); on Push-T the differences are $-6.0$, $+10.0$, and $-4.0$ (Supp.~Tabs.~S1 and S2).

Our planner also exceeds two-level H-JEPA search on all four environments (by 8.0--33.8 percentage points, and in all 12 backbone--environment pairs) and LeWM (by 2.0 percentage points on Push-T and by more elsewhere). H-JEPA also reports a three-level configuration (73.3\%$\pm$3.5 on AntMaze under its own protocol~\cite{zhang2026hjepa}); we compare against the two-level configuration, which matches our two-level backbone, and do not evaluate three-level H-JEPA under the common protocol. Subgoals generated from learned conditions yield higher task success than linear latent interpolation, which reaches at most 4.7\%.

\begin{figure*}[t]
\centering
\includegraphics[width=0.6\textwidth]{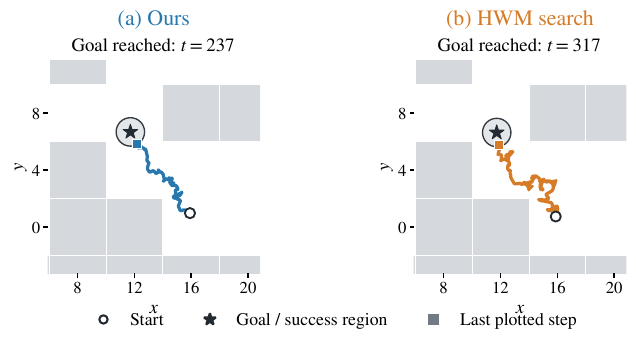}
\caption{\textbf{AntMaze navigation, task 0.} Recorded planar positions for our planner (a) and HWM search (b), using backbone 42 and head initialization 0. Axes use native simulator coordinates; gray regions are walls and shaded circles are goal regions. Realized start and goal positions differ slightly between panels because of stochastic resets (\cref{sec:protocol}). Paths end at first success, or at the 350-step budget on failure; panel titles give the first-success step. Aggregate success is reported in \cref{tab:main}.}
\label{fig:qual_ant}
\end{figure*}

All nine FourRoom head runs solve all 50 fixed tasks. \Cref{fig:qual_ant} illustrates the AntMaze behavior on task 0: our planner reaches the goal at $t=237$, and HWM search at $t=317$, from their respective realized initial states. \suppref{H} provides Cube and FourRoom examples.

\subsection{Computational cost}
\label{sec:cost}

\begin{figure*}[t]
\centering
\includegraphics[width=0.86\textwidth]{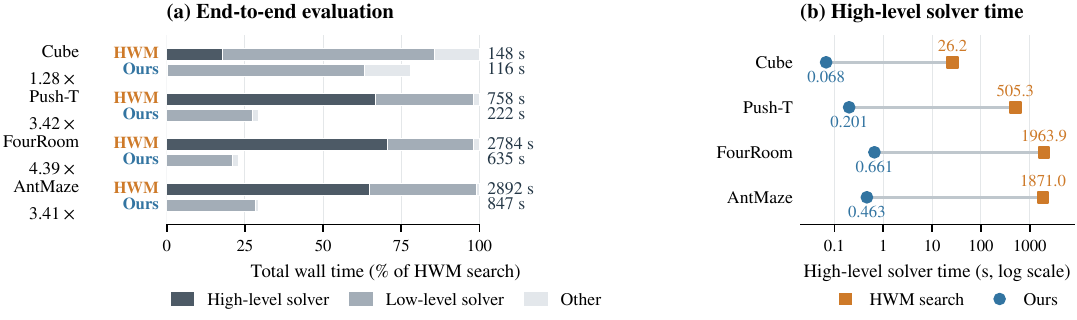}
\caption{\textbf{Batched evaluation time on one L40S GPU.} Each measurement evaluates 50 tasks in parallel with backbone 42 and head initialization 0; there is one paired measurement per environment. (a) Paired wall times normalized to HWM search within each environment; labels give absolute seconds and end-to-end speedup. ``Other'' is the time outside the two solvers, including observation processing and simulation. (b) Accumulated high-level solver time on a logarithmic axis. Timing uses a different batching protocol from success evaluation and the released matching weight $\beta=0.2$ for HWM search on Cube and Push-T; see \cref{sec:cost}.}
\label{fig:timing}
\end{figure*}

Replacing candidate optimization with $H$ head--predictor calls reduces high-level solver time by more than two orders of magnitude (\cref{fig:timing}b), leaving the retained low-level planner as the dominant cost; end to end, evaluation is \mbox{$1.28$--$4.39\times$} faster than HWM search. We time one backbone--head pair (backbone 42, head 0), selected before inspecting results, against its HWM search baseline on a dedicated NVIDIA L40S GPU, synchronizing the device before each timer read. Timing measures the wall time of one parallel batch of 50 environments, whereas success comparisons use the chunk-one protocol. For Cube and Push-T, the timing baseline keeps the released matching weight $\beta=0.2$, whereas the head uses zero, so the measured speedups compare these planner configurations, including their different low-level objectives.

High-level solver time per batch falls from \mbox{26.2--1{,}963.9~s} to \mbox{0.068--0.661~s} (\cref{fig:timing}). Low-level solver time is also 7--24\% lower: different subgoals change trajectories and early-stopping points, and Cube and Push-T additionally use different matching weights.

Across nine head runs per environment, median training time is 6.9~min per head on Cube, 32.2~min on Push-T, 24.0~min on FourRoom, and 26.0~min on AntMaze. These L40S measurements include optimization and periodic validation but exclude embedding-cache preparation; cache preparation and head training are one-time costs. \supprefs{C and F} report training-time ranges and complete evaluation timings.

\subsection{Ablation studies}
\label{sec:ablation}

We train each ablation at backbone seed 42 with three head initializations, the same training samples for a given segment length, and the same validation windows. \Cref{tab:ablation} reports means and sample standard deviations (SD) across these three runs; Supp.~Tabs.~S6 and S7 list every variant.

\begin{table*}[t]
\caption{\textbf{Condition-parameterization and head-input ablations (backbone 42, three heads).} Success (\%), mean $\pm$ sample SD over three head runs. The ``Ours (full)'' row repeats the backbone-42 rows of Supp.~Tab.~S1; the variants are defined in \suppref{E}. A dash denotes a configuration that was not run.}
\label{tab:ablation}
\centering\small\setlength{\tabcolsep}{8pt}
\input{tables/main_ablation.tex}
\end{table*}

\paragraph{Routing subgoals through the predictor.}
The direct variant regresses the next latent state without $F_2$ on the same hindsight data, a latent-subgoal regression in the spirit of the high-level policy of HIQL~\cite{park2023hiql} but without its advantage weighting. Direct regression has higher means on Cube and Push-T and lower means on FourRoom and AntMaze (\cref{tab:ablation}); the largest difference is on AntMaze, where the mean is 70.7\% with subgoals routed through $F_2$ versus 59.3\% with direct regression. We retain the predictor-mediated form for the reason given in \cref{sec:method}; at three head initializations this single-set difference is suggestive rather than conclusive (\suppref{E.1}).

\paragraph{Conditioning on the current state.}
Masking the head's state input reduces mean success to 52.0\%, 0.7\%, 6.7\%, and 2.0\% on the four sets (\cref{tab:ablation}). Because $F_2$ still receives the latent state context, the drop isolates the contribution of a state-dependent condition.

\paragraph{Four-dimensional versus unrestricted conditions.}
We train a head that outputs a four-dimensional macro-action, then apply the frozen lift $A$ before $F_2$. The two parameterizations reach broadly similar mean success (\cref{tab:ablation}); we report the unrestricted one because it keeps the head independent of the macro-action pathway (\cref{sec:method}), not because it improves success. Restricted to the four-dimensional space the head still reaches 73.3\% against 48.0\% for search at backbone 42, and the condition-space geometry of \suppref{G} does not translate into higher success (\suppref{E.1}).

\paragraph{Losses and supervision.}
Removing the rollout term lowers mean success on Push-T from 44.0\% to 33.3\%, whereas on Cube the mean increases from 68.7\% to 71.3\%. The teacher-forced and rollout losses share their first-transition error, and the endpoint error also appears in the final step of the rollout loss; these overlaps help interpret the ablations but do not establish that the other terms are unnecessary (\suppref{E.1}). Supervising the head with action-derived condition labels reaches 24.7\% on Push-T, compared with 44.0\% for the full method. This comparison favors training through the frozen predictor in this setting, but does not by itself establish why label supervision performs worse. Segment length is the most sensitive setting and also changes the eligible training trajectories (\suppref{C}); the full sweep is in Supp.~Tab.~S6.

\section{Discussion and Limitations}
\label{sec:limitations}

\paragraph{Implications for control.}
The results support learning high-level conditions to adapt a pretrained predictive hierarchy for goal-directed control. Intermediate-state supervision uses transitions realized during execution as references for plausible state changes, while the endpoint objective emphasizes proximity to the selected goal. The state-input ablation supports making these conditions depend on the current state (\cref{tab:ablation}).

\paragraph{Scope of the results.}
Results are measured on four fixed simulated benchmark sets with the released shared-latent HWM backbones. The method requires a differentiable high-level predictor whose outputs lie in the low-level planner's target space. Low-level action search remains the main deployment cost, and end-to-end savings depend on hardware, batching, and solver settings. Embedding-cache preparation and head training add one-time costs.

\section{Conclusion}

We introduced a condition head that supplies goal-dependent vectors to HWM's frozen high-level predictor to generate subgoals for the retained low-level action planner, eliminating repeated high-level candidate macro-action optimization. Only the head is trained, using recorded intermediate states and endpoints to encourage consistency with observed transitions and proximity to the goal. On the four fixed benchmark sets, mean task success is 19--36 percentage points higher than HWM search on three sets and equal on the fourth; batched timing runs show \mbox{$1.28$--$4.39\times$} end-to-end speedups.

%% file: tables/main_results.tex
\begin{tabular}{lcc@{\hskip 14pt}ccc}
\toprule
& \multicolumn{2}{c}{Separately trained} & \multicolumn{3}{c}{Same frozen HWM backbone} \\
\cmidrule(lr){2-3}\cmidrule(lr){4-6}
Environment & H-JEPA (two-level) & LeWM (single-level) & HWM search & Linear waypoints & Ours \\
\midrule
Cube & $46.7\,{\scriptstyle\pm\,6.7}$ & $26.0\,{\scriptstyle\pm\,1.2}$ & $51.3\,{\scriptstyle\pm\,5.5}$ & $4.7\,{\scriptstyle\pm\,1.8}$ & $\mathbf{70.4}\,{\scriptstyle\pm\,2.5}$ \\
Push-T & $40.7\,{\scriptstyle\pm\,5.9}$ & $46.7\,{\scriptstyle\pm\,2.9}$ & $\mathbf{48.7}\,{\scriptstyle\pm\,2.4}$ & $1.3\,{\scriptstyle\pm\,0.7}$ & $\mathbf{48.7}\,{\scriptstyle\pm\,2.9}$ \\
FourRoom & $78.7\,{\scriptstyle\pm\,6.4}$ & $40.0\,{\scriptstyle\pm\,3.1}$ & $72.0\,{\scriptstyle\pm\,7.0}$ & $0.0\,{\scriptstyle\pm\,0.0}$ & $\mathbf{100.0}\,{\scriptstyle\pm\,0.0}$ \\
AntMaze & $37.3\,{\scriptstyle\pm\,1.8}$ & $18.0\,{\scriptstyle\pm\,3.1}$ & $35.3\,{\scriptstyle\pm\,0.7}$ & $0.0\,{\scriptstyle\pm\,0.0}$ & $\mathbf{71.1}\,{\scriptstyle\pm\,1.9}$ \\
\bottomrule
\end{tabular}

%% file: tables/main_ablation.tex
\begin{tabular}{lcccc}
\toprule
Variant & Cube & Push-T & FourRoom & AntMaze \\
\midrule
Ours (full) & $68.7\,{\scriptstyle\pm\,4.2}$ & $44.0\,{\scriptstyle\pm\,2.0}$ & $100.0\,{\scriptstyle\pm\,0.0}$ & $70.7\,{\scriptstyle\pm\,6.1}$ \\
Direct subgoals & $69.3\,{\scriptstyle\pm\,4.2}$ & $46.0\,{\scriptstyle\pm\,4.0}$ & $98.7\,{\scriptstyle\pm\,1.2}$ & $59.3\,{\scriptstyle\pm\,4.2}$ \\
No head state input & $52.0\,{\scriptstyle\pm\,3.5}$ & $0.7\,{\scriptstyle\pm\,1.2}$ & $6.7\,{\scriptstyle\pm\,3.1}$ & $2.0\,{\scriptstyle\pm\,2.0}$ \\
4D conditions via $A$ & $73.3\,{\scriptstyle\pm\,1.2}$ & $40.7\,{\scriptstyle\pm\,8.3}$ & -- & $70.7\,{\scriptstyle\pm\,5.0}$ \\
\bottomrule
\end{tabular}

%% file: AI_ASSISTANCE_DISCLOSURE.tex
OpenAI Codex assisted with drafting and revising the Abstract, Sections 1--6, and Supplementary Sections A--J; experiment planning and analysis; developing and debugging experiment and analysis scripts; and preparing diagram and plotting code for Figures 1--4 and Supplementary Figure S1. Anthropic Claude assisted with manuscript review and revision planning, revised the text of the main paper and supplement, and prepared the TikZ sources of Figures 1--2 and the plotting scripts of Figures 3--4. Quantitative results and qualitative trajectories were obtained from recorded experimental runs. The authors retain responsibility for the scientific content, interpretation, and conclusions.

%% file: supplement.tex
\setcounter{section}{0}
\setcounter{table}{0}
\setcounter{figure}{0}
\setcounter{equation}{0}
\crefname{appendix}{Sec.}{Secs.}
\Crefname{appendix}{Section}{Sections}
\renewcommand{\thesection}{\Alph{section}}
\renewcommand{\thetable}{S\arabic{table}}
\renewcommand{\thefigure}{S\arabic{figure}}
\renewcommand{\theequation}{S\arabic{equation}}
\renewcommand{\theHsection}{supp.\Alph{section}}
\renewcommand{\theHtable}{supp.\arabic{table}}
\renewcommand{\theHfigure}{supp.\arabic{figure}}
\renewcommand{\theHequation}{supp.\arabic{equation}}
\setcounter{dbltopnumber}{4}
\renewcommand{\dbltopfraction}{0.92}
\renewcommand{\textfraction}{0.07}
\renewcommand{\dblfloatpagefraction}{0.80}
\setlength{\dblfloatsep}{12pt plus 2pt minus 2pt}
\setlength{\dbltextfloatsep}{14pt plus 2pt minus 2pt}
\raggedbottom
\pdfbookmark[0]{Supplementary Material}{supplementary-material}
\twocolumn[\centering{\Large\bfseries Supplementary Material\par}\vspace{16pt}]

\noindent\textbf{Overview.} This supplement provides per-run results and the statistics behind the main tables (\cref{sec:supp_protocol}, \cref{tab:heads,tab:baselines}); data, normalization, and model selection (\cref{sec:data}); architecture and training details with training cost (\cref{sec:implementation}, \cref{tab:architecture,tab:training}); inference, schedules, and baseline definitions (\cref{sec:supp_inference}, \cref{tab:solver_settings}); the complete ablation tables (\cref{sec:ablations}, \cref{tab:cubepushabl,tab:fourantabl}); dedicated evaluation timing (\cref{sec:timing}, \cref{tab:timing}); a linear condition-space diagnostic (\cref{sec:pca}); qualitative rollouts (\cref{sec:qual}, \cref{fig:qual_pair}); reproduction details (\cref{sec:repro}); and additional evaluations with disjoint source groups (\cref{sec:group_evaluation}, \cref{tab:group_partitions,tab:group_results}). Notation follows the main paper.

\section{Evaluation protocol and statistics}
\label{sec:supp_protocol}

The benchmark evaluations in \cref{tab:heads,tab:baselines} use the same 50 tasks for each environment, in the order specified by the provided task file. Evaluation chunk size is one for every method, i.e., one task is evaluated at a time with unbatched solver calls. Backbone seeds are 42, 43, and 44; each task's planner random stream is seeded by the evaluator's per-task rule. Our planner has head initializations 0, 1, and 2 for each backbone; the segment-sampling rule that pairs runs of equal training length is given in \cref{sec:implementation}.

\Cref{tab:heads,tab:baselines} report the individual head and baseline runs. State-prediction and validation errors use the mean absolute error (MAE) across latent coordinates, as defined in the main paper.

Let $b\in\{42,43,44\}$ index the backbone seed, $h\in\{0,1,2\}$ the head initialization, and $s_{b,h}$ the percentage of successful tasks for that run. Denote the within-backbone mean by $\bar s_b$, the overall mean by $\bar s$, and the standard error across backbone means by $\mathrm{SE}_{\mathrm{backbone}}$. With $\mathrm{SD}$ denoting the sample standard deviation (Bessel-corrected, $n-1=2$ for three values), these statistics are
\begin{align}
 \bar s_b&=\frac13\sum_{h=0}^2s_{b,h},\qquad
 \bar s=\frac13\sum_{b\in\{42,43,44\}}\bar s_b,\\
 \mathrm{SE}_{\mathrm{backbone}}&=
 \frac{\mathrm{SD}(\bar s_{42},\bar s_{43},\bar s_{44})}{\sqrt3}.
\end{align}
Baselines have one run per backbone and report the analogous standard error across three values. The backbone mean is the statistical unit, and all runs share the fixed task set. Our planner has zero standard error on FourRoom because all nine heads solve all 50 tasks.

Ablations report the mean and sample standard deviation of three head runs at backbone 42. The released task assets specify task identities while leaving some reset randomness, including FourRoom's distractor behavior, to the environment. Run spread therefore reflects head initialization, GPU nondeterminism, and environment randomness.

\begin{table*}[t]
\caption{\textbf{Our planner: per-head success (\%).} All heads complete 20{,}000 training updates and are selected by validation endpoint MAE (\cref{sec:data}). Each row is one fixed backbone.}
\label{tab:heads}
\centering\small\setlength{\tabcolsep}{10pt}
\input{tables/per_head.tex}
\end{table*}

\begin{table*}[t]
\caption{\textbf{Baseline success (\%).} Every listed cell uses chunk size one. HWM and two-level H-JEPA search both use matching weight $\beta=0$. LeWM is a single-level planner; LeWM and H-JEPA have their own pretrained weights, and only HWM search shares our planner's backbone. The linear-waypoints baseline uses the same frozen HWM encoder and low-level planner.}
\label{tab:baselines}
\centering\small\setlength{\tabcolsep}{10pt}
\input{tables/per_baseline.tex}
\end{table*}

\section{Data, normalization, and model selection}
\label{sec:data}

\paragraph{Trajectory datasets.}
The benchmark datasets contain 9{,}800 training and 200 validation trajectories for Cube, 18{,}685 and 21 for Push-T, and 25{,}000 and 1{,}250 for AntMaze. FourRoom uses 3{,}750 training trajectories and 150 validation trajectories collected separately. Embedding caches retain trajectory boundaries, and sampled segments never cross between episodes.

\paragraph{FourRoom validation.}
Training collection uses seed 3072 and validation collection uses seed 1003072. The separately collected validation trajectories select condition-head checkpoints. The pretrained backbone checkpoints are inherited from the release.

\paragraph{Benchmark tasks.}
Cube and Push-T use the released 50-task assets \texttt{goal\_offset\_20\_pickup\_val.pt} and \texttt{goal\_offset\_75\_val.pt} (whose goals lie 74 environment steps after the start frame), respectively, constructed from their public validation trajectories. The same validation corpora supply the windows used for checkpoint selection. FourRoom and AntMaze use the provided navigation task assets. Success is determined by each environment's released success check and recorded if it is reported at any executed step (\cref{sec:supp_inference}).

\paragraph{AntMaze inputs.}
The observation encoder combines a 256-dimensional visual embedding with a 128-dimensional proprioceptive embedding, giving latent dimension $d=384$. Proprioception is formed by concatenating \texttt{qpos[2:]} and \texttt{qvel}; both cache generation and control apply the checkpoint's evaluation mean and standard deviation before the proprioceptive encoder. Pixel preprocessing is identical in the two paths. All AntMaze heads, ablations, and head timing measurements reported here use this input convention.

\paragraph{Checkpoint selection.}
We draw 256 validation windows using NumPy random seed 7 and reuse exactly those windows at every validation check, every 2{,}000 updates through update 20{,}000. The checkpoint with minimum autoregressive endpoint MAE on this fixed subset is selected; ties retain the earlier checkpoint. Task success is evaluated after checkpoint selection and is never used for selection. Let $K_{\mathrm{val}}$ be the number of transitions in each validation window. All variants, including short- and long-segment training, use $K_{\mathrm{val}}$ equal to the environment's default $K$ and the same windows. Validation uses recorded endpoints as goals and the training rule $r_{k,j}=K_{\mathrm{val}}-j$ for the remaining-step inputs (ending at 1); the terminal-zero deployment convention is specified in \cref{sec:supp_inference}. Validation predictions are autoregressive, with the head in evaluation mode and gradients disabled. The main comparison aggregates all specified backbone and head initializations using the statistics in \cref{sec:supp_protocol}.

\section{Architecture and training details}
\label{sec:implementation}

\Cref{tab:architecture} lists the training horizons and predictor sizes. We use the main paper's notation: $\bm z\in\mathbb R^d$ is a latent state, $\bm z_g\in\mathbb R^d$ the goal, $K$ the training segment length in transitions, and $F_2$ the frozen coarse predictor. Let $r$ be a nonnegative integer remaining-step input. For frequency indices $m=0,\ldots,7$, define $\omega_m=2^m\pi/16$. We write $[\cdot;\cdot]$ for vector concatenation. The 16-dimensional sinusoidal positional encoding (PE) of the remaining-step input is
\begin{equation}
 \operatorname{PE}(r)=\big[(\sin(\omega_m r))_{m=0}^7;\,(\cos(\omega_m r))_{m=0}^7\big].
\end{equation}
The condition head receives the concatenation of state, goal, and this encoding:
\begin{equation}
 [\bm z;\bm z_g;\operatorname{PE}(r)]\in\mathbb R^{2d+16}.
\end{equation}
The network applies layer normalization, a linear layer from $2d+16$ to 512 units, a Gaussian error linear unit (GELU) activation, another 512-unit linear layer and GELU, then a linear output layer with $d$ units. Head parameter counts are 567{,}264 for $d=192$ and 863{,}136 for $d=384$. The predictor context contains at most three state/condition pairs, ordered in time; early steps use the available shorter context. During a high-level rollout, later context states are generated latents.

\begin{table}[t]
\caption{\textbf{Training horizon and frozen predictor size.} $K$ counts transitions, while a segment has $K+1$ states. Frozen coarse-predictor ($F_2$) parameter counts are shown in millions, rounded to 0.01\,M.}
\label{tab:architecture}
\centering\small\setlength{\tabcolsep}{5pt}
\begin{tabular}{lrrrr}
\toprule
Environment & $d$ & $K$ & States & $F_2$ params (M)\\
\midrule
Cube & 192 & 2 & 3 & 11.58\\
Push-T & 192 & 8 & 9 & 11.58\\
FourRoom & 192 & 12 & 13 & 4.84\\
AntMaze & 384 & 9 & 10 & 10.25\\
\bottomrule
\end{tabular}
\end{table}

The head uses AdamW with learning rate $10^{-3}$, weight decay $10^{-4}$, cosine decay over 20{,}000 updates, batch size 256, and gradient-norm clipping at 1.0. The additional endpoint-loss weight $\lambda$ is one unless ablated. The default width (512), endpoint weight, learning rate, and 20{,}000-update budget were fixed before inspecting experimental results. The backbone and normalizer remain fixed during head training. Gradients propagate through the frozen predictor to update the condition head, so training includes both forward and backward computation through the predictor.

Training samples an eligible trajectory uniformly, then a valid start uniformly within it, using a NumPy generator initialized with seed 0. Thus equal-length runs share the segment-sampling sequence across head initializations. In the fine-level embedding cache (stride 5 environment steps), a $K$-transition segment uses $2K+1$ cache entries and selects every second entry. Eligibility requires an episode long enough for the entire segment. For the default Push-T length ($K=8$), 16{,}463 of the 18{,}685 training trajectories are eligible; doubling the training horizon reduces the eligible pool to 2{,}727 trajectories, changing both the temporal scale and the sampled training population.

Training uses the three MAE terms defined in the main paper. Teacher forcing supervises local predictions from recorded contexts; rollout supervision compares recursively generated states with the recorded sequence; endpoint supervision adds weight to the hindsight goal. Intermediate targets supply feasibility cues from state changes realized during execution. They supervise states in the encoder's latent space, while the low-level planner chooses primitive actions at deployment. The autoregressive endpoint has total weight $1/K+\lambda$, including its contribution inside the rollout loss. An update makes $2K$ batched coarse-predictor calls: $K$ teacher-forced and $K$ autoregressive. Encoder evaluation occurs during cache preparation.

\paragraph{Training cost.}
\Cref{tab:training} summarizes nine L40S training runs per environment, including optimization and validation but excluding cache loading and preparation. Let $T_{\mathrm{cache}}$ and $T_{\mathrm{train}}$ denote one-time cache-preparation and head-training wall times, and $T_{\mathrm{online}}$ the wall time per evaluation at fixed batch size and hardware; the total time for $M$ evaluations is $T_{\mathrm{cache}}+T_{\mathrm{train}}+M\,T_{\mathrm{online}}$. We report head-training and evaluation times separately; cache preparation lies outside both measurements.

\begin{table}[t]
\caption{\textbf{Per-head training wall time.} Each run performs 20{,}000 updates. Nine runs are summarized per environment; values are minutes. Cache preparation is excluded.}
\label{tab:training}
\centering\small\setlength{\tabcolsep}{4pt}
\input{tables/training_cost.tex}
\end{table}

\section{Inference details and baseline definitions}
\label{sec:supp_inference}

\paragraph{High-level remaining-step inputs and rollout.}
For an active high-level rollout with horizon $H>1$, let $k=0,\ldots,H-1$ be the prediction-call index and $j\in\mathcal I_k=\{\max(0,k-2),\ldots,k\}$ a retained context position. The remaining-step input at that position is $r_{k,j}=H-j$ for $k<H-1$, and zero for all positions when $k=H-1$, so the inputs attached to the most recent state are $H,H-1,\ldots,2,0$, whereas training uses $r_{k,j}=K-j$ with inputs $K,\ldots,1$. Both head and predictor are evaluated $H$ times, giving $H+1$ states including the initial state; each head evaluation computes the conditions of all retained context positions in one batch, and the terminal evaluation recomputes them with the zero inputs. Each replanning call starts a new latent rollout from the current observation; previous predicted contexts are discarded. Only the first predicted state $\hat{\bm z}_1=F_2(\bm z_t,g_\theta(\bm z_t,\bm z_g,H))$ is passed to the low level; it is computed at $k=0$ from the single observed state with remaining-step input $H$, so the terminal-zero convention and the later context states do not alter the selected subgoal. The later states are computed by the retained rollout implementation, and the timing in \cref{sec:timing} includes them. \main{Algorithm~\ref{alg:plan}} states the inference rule.

\paragraph{Horizon and execution schedule.}
Let $B$ be the budget in environment steps, $t$ the number already executed, and $n=\lceil B/10\rceil$ the number of nominal schedule positions. For an initial horizon $H_0$, define the descending integer sequence $S(n,H_0)$ of length $n$, for $n\geq H_0$ (true for all schedules used here). With $q=\lfloor n/H_0\rfloor$ and remainder $\rho=n-qH_0$, each value from $H_0$ down to one appears $q$ times, and each of the smallest $\rho$ values appears once more. The high-level schedule is $(H_i)_{i=0}^{n-1}=S(n,H_0)$. Budgets $B$ are 30, 90, 210, and 350, and $H_0$ values are 2, 8, 12, and 9 for Cube, Push-T, FourRoom, and AntMaze, respectively.

Let $i_0$ be the first schedule position with $H_{i_0}=1$. The low-level horizon $L_i$ counts blocks of five independently optimized primitive actions; $L_i=2$ for $i<i_0$, and from $i_0$ onward the tail is $S(n-i_0,L_{\mathrm{tail}})$, where $L_{\mathrm{tail}}$ is given in \cref{tab:solver_settings}. At execution time, the schedule index is $i(t)=\lfloor t/10\rfloor$, so $H(t)=H_{i(t)}$ and $L(t)=L_{i(t)}$. The controller executes $\min(10,5L(t),B-t)$ primitive actions before replanning. Consequently, a one-block plan triggers a new solve after five steps and may reuse the same schedule position. For example, Cube uses $H_i=(2,1,1)$ and $L_i=(2,2,1)$, with solves at $t=0,10,20,25$.

Evaluation runs for the full budget, records whether success occurred at any executed step, and suppresses automatic environment reset during the rollout. Success does not terminate the evaluation loop. Qualitative plots are truncated at first success for display, whereas evaluation wall times include the full budget.

\paragraph{Low-level objective and action selection.}
Let $L=L(t)$, and let $\bm z_{\mathrm{tar}}$ denote the subgoal $\hat{\bm z}_1$ when $H(t)>1$, or the encoded task goal when $H(t)=1$. For a candidate primitive-action sequence $\bm u_{t:t+5L-1}$, let $\bm v_m$ be the state predicted through $F_1$ after $m$ five-action blocks, with $\bm v_0=\bm z_t$. The planning horizon $\ell$ in \main{Fig.~\ref{fig:overview}} is $5L$ environment steps, and the figure's $\hat{\bm z}_{t+\ell}$ is $\bm v_L$. Define the scored block indices as $\mathcal B_t=\{2\}$ while a high-level subgoal is active, in which case $L=2$, and $\mathcal B_t=\{\max(1,L-1),\ldots,L\}$ when the high level is bypassed. In the shared-latent backbone, the low-level objective is
\begin{equation}
 J_{\mathrm{low}}(\bm u_{t:t+5L-1})
 =\sum_{m\in\mathcal B_t}\|\bm v_m-\bm z_{\mathrm{tar}}\|_2^2.
\end{equation}
The subgoal case scores the ten-step prediction aligned with the coarse target; the goal case scores up to the last two fine predictions. The low-level solver differentiates through $F_1$ to update candidate actions, keeps all model parameters fixed, and returns the candidate with the lowest final cost. The policy converts normalized actions back to environment units before execution.

\begin{table}[t]
\caption{\textbf{Retained solver settings.} $L_{\mathrm{tail}}$ is the configured low-level tail horizon in five-action blocks, $N_{\mathrm{low}}$ the number of low-level candidates, and $\eta_{\mathrm{low}}$ and $\eta_{\mathrm{high}}$ the AdamW learning rates for low-level planning and high-level baseline search. High-level candidate counts are in \main{Tab.~\ref{tab:budget}}.}
\label{tab:solver_settings}
\centering\small\setlength{\tabcolsep}{3pt}
\begin{tabular}{lrrrr}
\toprule
Environment & $L_{\mathrm{tail}}$ & $N_{\mathrm{low}}$ & $\eta_{\mathrm{low}}$ & $\eta_{\mathrm{high}}$\\
\midrule
Cube & 2 & 400 & 0.10 & 0.075\\
Push-T & 2 & 50 & 0.10 & 0.075\\
FourRoom & 2 & 100 & 0.08 & 0.10\\
AntMaze & 4 & 25 & 0.06 & 0.10\\
\bottomrule
\end{tabular}
\end{table}

Both the high-level search and the low-level solver allow at most 90 updates. Early stopping is allowed after 30 updates and requires five consecutive updates without a relative improvement exceeding 1\% in the tracked best score. The score is the mean, over the solver batch, of each task's lowest candidate cost. Each solve initializes one candidate at the action-prior mean and the others with Gaussian perturbations. When stored training actions are available, their coordinate means and standard deviations define this prior and their 2nd/98th percentiles define clipping bounds after updates; otherwise initialization uses zero mean and unit standard deviation in normalized action coordinates. On Cube and Push-T, the high-level search perturbs the candidate macro-actions with Gaussian noise of standard deviation 0.01 after each update, as in the released configuration; the FourRoom and AntMaze searches and the low-level solver add no noise. \Cref{tab:solver_settings} gives the learning rates; other AdamW options retain the defaults of the software version specified in \cref{sec:repro}.

\paragraph{Linear waypoints.}
For the linear baseline, let $\hat{\bm z}_0=\bm z_t$ be the current observed latent and $\hat{\bm z}_k$ its $k$th waypoint. With goal $\bm z_g$ and active horizon $H$, it predicts
\begin{equation}
 \hat{\bm z}_k=\hat{\bm z}_0+\frac{k}{H}(\bm z_g-\hat{\bm z}_0),
 \quad k=1,\ldots,H.
\end{equation}
It hands the first waypoint to the same low-level planner and replans using the current observation and active horizon. In the context-based implementation, the latest predicted state $\bm z$ and its remaining-step input $r$ give the next state as $\bm z+(\bm z_g-\bm z)/\max(r,1)$; the terminal zero input yields the goal. This generates equally spaced waypoints over the full active horizon, without a learned dynamics model, trainable head, or high-level search.

\paragraph{Matching weight $\beta$.}
The released low-level controller includes an auxiliary loss that matches its predicted action embedding to the high-level macro-action. Its default weight is $\beta=0.2$ on Cube and Push-T, and zero on FourRoom and AntMaze. Our head supplies a condition directly in predictor-input space, so $\beta$ is zero. Both hierarchical search baselines also use zero in the main success comparison, a deviation from the released Cube and Push-T setting. The separate timing comparison uses the weights stated in \cref{tab:timing}.

\paragraph{Baseline optimization.}
The original HWM paper defines L1 planning costs and reports the cross-entropy method for Franka and Push-T and model predictive path integral control for Diverse Maze (Sec.~2.2 and App.~B of the arXiv version of the HWM paper, arXiv:2604.03208). We instead evaluate the HWM configuration distributed with H-JEPA, which uses gradient-based search and a squared L2 terminal cost. High-level search uses the budgets in \main{Tab.~\ref{tab:budget}}, with the update limit, initialization, clipping, noise, and early-stopping rule specified above. Its objective is the squared L2 distance between the terminal predicted state and the goal (\main{Eq.~(\ref{eq:search})}); only the final state is scored at the high level. Macro-action priors and clipping bounds come from stored training macro-actions. The search rollout uses its released one-state context, whereas learned-condition rollouts use up to three state/condition pairs. Comparisons use these fixed search configurations across backbone seeds. The H-JEPA baseline is the two-level configuration, run with the same high-level budgets, learning rates, and noise settings as HWM search in each environment; H-JEPA's three-level configuration is reported in its own paper under its own protocol and is not evaluated here. LeWM plans with its released single-level gradient solver: 800, 800, 200, and 100 candidates, 90 updates, horizons of 4, 15, 23, and 23 five-action blocks, and learning rates 0.05, 0.05, 0.10, and 0.30 on Cube, Push-T, FourRoom, and AntMaze, respectively, with no optimization noise.

\section{Detailed ablation results}
\label{sec:ablations}

\Cref{tab:cubepushabl,tab:fourantabl} report the evaluated variants for Cube and Push-T, and for FourRoom and AntMaze, respectively. All variants fix backbone seed 42. Each head is trained with initializations 0, 1, and 2, and selected independently by the common validation criterion. The ``Ours (full)'' rows repeat the backbone-42 rows of \cref{tab:heads}. Sample standard deviations describe run-to-run variation within this backbone. The full loss and input sweep was run on Cube and Push-T; FourRoom and AntMaze received the segment-length, condition-parameterization, and state-input variants listed, and the FourRoom four-dimensional (4D) configuration was not run.

\paragraph{Predictor-mediated vs.\ direct subgoals.}
The direct head has the same inputs and hidden architecture as the full head, but its output is the next latent state. Teacher-forced, rollout, and endpoint losses are computed directly on that output; $F_2$ is bypassed during its state prediction. The 4D variant outputs a macro-action and applies the frozen lift $A$ before $F_2$. Both variants train their head parameters with the same three state-prediction losses.

\paragraph{Head inputs.}
The no-state variant masks the head's state input during teacher forcing, autoregressive training, and deployment, while $F_2$ continues to receive the recorded or predicted latent context. The no-remaining-step variant sets the head's remaining-step input to zero during both training and deployment.

\paragraph{Loss and supervision variants.}
Each loss-removal variant sets the corresponding objective coefficient to zero. The teacher-forcing-only variant sets both rollout and endpoint coefficients to zero. Removing the separate endpoint loss leaves the endpoint error inside the rollout loss. Removing the rollout loss retains the autoregressive endpoint term, so that variant still backpropagates through a generated trajectory. Endpoint weights 0.1 and 10 change only $\lambda$. All variants use the same autoregressive endpoint validation criterion. In the ``Supervised lifted conditions'' variant, the target is $A(\bm a)$, where $\bm a\in\mathbb R^4$ is a macro-action inferred by the frozen action encoder from recorded action sequences and $A$ is the frozen macro-action lift. In the ``Supervised macro-actions'' variant, the target is $\bm a$ itself, followed by $A$ at deployment. Both use MAE regression to action-derived labels. The main head objective supervises predicted states using recorded observation sequences.

\paragraph{Short and long segments.}
The short choice uses $\max(1,\lfloor K/2\rfloor)$ transitions, and the long choice uses $2K$. All lengths use the same default-$K$ validation windows for checkpoint selection. Training trajectories must accommodate the sampled segment, so changing its length also changes the eligible trajectory pool, as described in \cref{sec:implementation}. These experiments assess the combined effect of segment length and the resulting sampling population.

\subsection{Interpretation of the ablation results}
\label{supp:ablation-interp}

\paragraph{Overlap among the loss terms.}
The teacher-forced and rollout losses share their first-transition error, while the endpoint error is also included in the final step of the rollout loss. Removing teacher forcing changes mean success by approximately $-1.3$ and $+0.7$ percentage points on Cube and Push-T, respectively; removing the separate endpoint term changes it by $+1.3$ and $-3.3$ points. Teacher forcing alone reaches 70.7\% and 31.3\%, compared with 71.3\% and 33.3\% when only the rollout term is removed from the full objective. These results show limited benefit from some additional terms under the tested settings, but do not establish that the rollout loss alone is sufficient: that combination was not evaluated in this sweep.

\paragraph{Why most Cube variants coincide.}
Cube uses $K=2$ and makes one high-level decision per episode (\cref{tab:timing}). The teacher-forced and rollout terms then differ in a single transition, and the deployment rollout never reads a state produced by an earlier prediction of the head. The limited exposure to self-generated inputs during deployment is consistent with the small differences among several Cube variants, although this explanation is not directly tested.

\paragraph{Direct subgoals and the number of high-level decisions.}
High-level decisions per episode are 1, 7, 19, and 31 on Cube, Push-T, FourRoom, and AntMaze (\cref{tab:timing}), while direct regression differs from the full method by $+0.7$, $+2.0$, $-1.3$, and $-11.3$ percentage points. On FourRoom, the full method reaches 100.0\%, leaving no room for direct regression to improve on it. The larger gap on AntMaze is consistent with a possible role for error accumulation over self-generated contexts, but differences between environments prevent attributing the pattern to the number of decisions alone. We do not test this mechanism, and the AntMaze comparison uses three head initializations per variant, with sample standard deviations of 6.1 and 4.2.

\paragraph{Four-dimensional conditions compared with search in the same space.}
The restricted head optimizes within the same four-dimensional macro-action space that HWM search explores. At backbone 42 on Cube it reaches 73.3\% against 48.0\% for search (\cref{tab:baselines}) under the budget of Tab.~1 of the main paper. Thus, on this Cube setting, higher mean success than search is also obtained without using the unrestricted output space. This comparison supports the learned mapping as an alternative to per-instance search under the tested budget, but does not isolate the cause of the performance difference.

\paragraph{Scope of the geometric diagnostic.}
\Cref{sec:pca} reports that learned unrestricted conditions lie outside the principal-component subspace of action-derived conditions and give lower one-step state error (0.119 versus 0.188). Both are properties of the condition space and of single-step prediction; the ablation table shows that neither implies higher task success.

\paragraph{Summary.}
Several reduced objectives have similar or higher means than the full objective on Cube ($K=2$), whereas Push-T ($K=8$) is more sensitive to the rollout loss and the endpoint weight; Sec.~4.4 of the main paper discusses these variants.

\begin{table*}[t]
\caption{\textbf{Cube and Push-T ablations.} Success (\%) for three head initializations at backbone 42, with mean $\pm$ sample SD. Default training lengths are 2 (Cube) and 8 (Push-T) transitions; short and long lengths are 1 and 4 (Cube) and 4 and 16 (Push-T). The teacher-forced term coincides with the rollout term on its first transition, and the endpoint error is contained in the final step of the rollout term; see \cref{supp:ablation-interp} for interpretation.}
\label{tab:cubepushabl}
\centering\small\setlength{\tabcolsep}{6pt}
\input{tables/ablation_cube_pusht.tex}
\end{table*}

\begin{table*}[t]
\caption{\textbf{FourRoom and AntMaze ablations.} Success (\%) for three head initializations at backbone 42, with mean $\pm$ sample SD. FourRoom runs use the separately collected validation set for checkpoint selection; default, short, and long training lengths are 12, 6, and 24 transitions. AntMaze runs use the training length of 9 transitions. A dash marks a configuration that was not run. See \cref{supp:ablation-interp} for interpretation.}
\label{tab:fourantabl}
\centering\small\setlength{\tabcolsep}{6pt}
\input{tables/ablation_fourroom_ant.tex}
\end{table*}

\begin{table*}[t]
\caption{\textbf{Batched evaluation timing on L40S.} Units are seconds for a batched evaluation of 50 tasks. Speedup is the total evaluation wall time of HWM search divided by that of our planner. Each environment has one paired measurement. HWM search uses its released matching weight $\beta=0.2$ on Cube and Push-T, and our planner uses zero; both use zero on FourRoom and AntMaze. High-level times are given to 0.001\,s because the learned values are below 1\,s. The last two columns give the number of low- and high-level solver calls per method, each operating on the parallel batch.}
\label{tab:timing}
\centering\small\setlength{\tabcolsep}{6pt}
\input{tables/timing_v6.tex}
\end{table*}

\FloatBarrier
\section{Dedicated evaluation timing}
\label{sec:timing}

\Cref{tab:timing} gives synchronized timing for the backbone-42/head-0 model. Each pair runs on a dedicated NVIDIA L40S GPU and evaluates 50 environments in one parallel batch, with the same task set and execution budget for both methods. The backbone--head pair was selected before inspecting results. Timing measures the wall time of the full evaluation batch; task-success comparisons use the separate chunk-one protocol.

The high and low columns accumulate synchronized solver wall time at the respective level. Total evaluation time additionally includes observation processing, simulation, and other evaluation work. Different subgoals lead to different trajectories and hence different early-stopping points, so equal solver call counts can still yield different low-level times.

\paragraph{Solver calls.}
In the timed evaluations, both methods make the same number of solver calls: (low-level, high-level) $=(4,1)$, $(10,7)$, $(22,19)$, and $(36,31)$ on Cube, Push-T, FourRoom, and AntMaze (\cref{tab:timing}). A call operates on the parallel batch. The difference between the two counts follows from the horizon schedule: the high level is bypassed once $H(t)=1$. Removing candidate optimization leaves 2, 8, 12, or 9 head--predictor calls at the initial horizons, respectively, of which the first determines the subgoal. The low-level planner still performs gradient-based action optimization. Speedups therefore depend on the original search budget, the low-level workload, batch size, and hardware.

\section{A linear condition-space diagnostic}
\label{sec:pca}

For backbone 42 and head initialization 0 on AntMaze, we fit principal component analysis (PCA) to 20{,}000 action-derived training conditions $A(\bm a)$. After centering at their mean, we retain four principal components. We calibrate projection residuals on 10{,}000 separate action-derived validation conditions and evaluate 4{,}096 learned conditions from fixed validation segments, with sampling seed 314159.

The components explain 99.95\% of training-condition variance. Mean Euclidean projection residual is 0.386 for action-derived validation conditions and 49.923 for learned conditions; all sampled learned conditions exceed the former distribution's 95th percentile, 0.909. On these segments, mean one-step state MAE is 0.119 with learned conditions and 0.188 with action-derived conditions.

The projection residuals quantify departure from a fitted affine subspace of action-derived conditions. On the sampled segments, learned conditions have substantially larger residuals than action-derived conditions. This subspace is a post hoc analysis tool and is distinct from the 4D-condition ablation, in which the head outputs four-dimensional macro-actions that pass through the frozen lift $A$; in that ablation, the 4D and unrestricted heads have the same mean AntMaze success. The diagnostic thus describes condition-space geometry, while the ablation compares control performance.

\clearpage
\onecolumn
\section{Qualitative rollouts}
\label{sec:qual}

\begin{figure}[!ht]
\centering
{\small\textbf{(a) Cube}}\\[-1pt]
\includegraphics[width=0.98\textwidth]{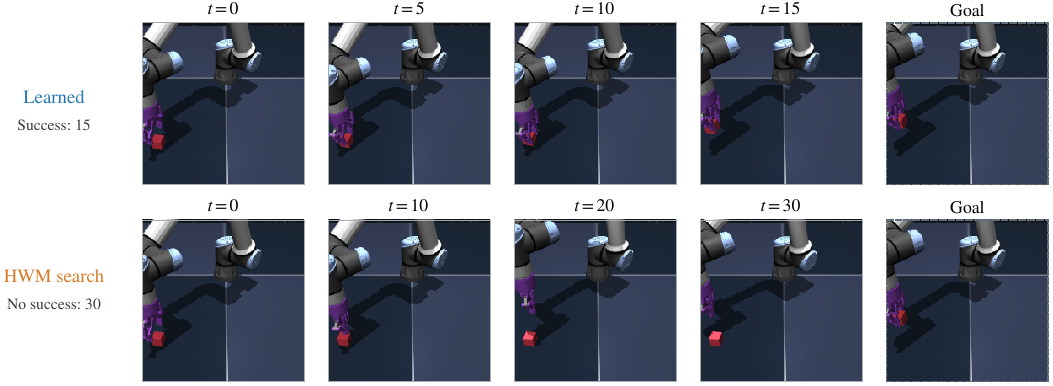}\\[4pt]
{\small\textbf{(b) FourRoom}}\\[-1pt]
\includegraphics[width=0.98\textwidth]{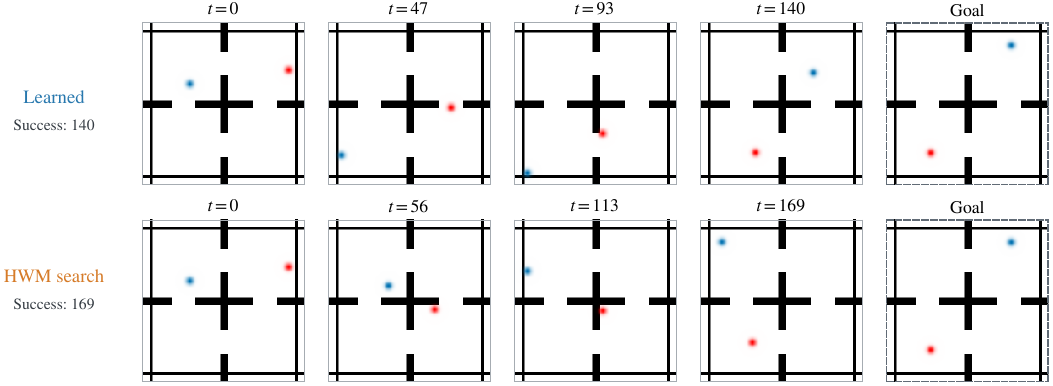}
\caption{\textbf{Qualitative rollouts on task 0.} Four observations per row are sampled uniformly through first success, or through the full budget on failure; each has its own recorded environment step $t$. The dashed final column is the task goal observation. (a) Cube: our planner first succeeds at step 15; HWM search does not succeed within the 30-step budget. (b) FourRoom: our planner and HWM search first succeed at steps 140 and 169. In (b), red is the controlled agent and blue a moving distractor that need not match the goal image.}
\label{fig:qual_pair}
\end{figure}

\Cref{fig:qual_pair} compares our planner (upper row of each panel, labeled ``Learned'') with HWM search (lower row) on task 0 of Cube and FourRoom. Both use backbone 42; the learned head uses initialization 0. Task 0 was selected before inspecting results, as was the AntMaze task-0 example in the main paper. These single rollouts illustrate behavior; aggregate success rates are given in \main{Tab.~\ref{tab:main}}.

\section{Reproduction details}
\label{sec:repro}

The experiments use Python 3.11.13, PyTorch 2.7.1, and CUDA 12.8. The model source is based on commit \texttt{840e76b5} of the H-JEPA codebase with the goal-conditioned rollout implementation accompanying the model package. Encoder embeddings, action lifts, normalizers, and predictors are fixed. Task-success runs retain the original reset semantics and per-task planner seeds, with chunk size one and the common matching weight $\beta=0$ described above.

We will publicly release the experiment code and trained condition-head checkpoints upon acceptance.

\clearpage
\twocolumn
\section{Additional evaluations with disjoint source groups}
\label{sec:group_evaluation}

We additionally compare HWM search and our planner on Cube and Push-T using separate training, head-checkpoint-selection, and test groups drawn from the released training corpora in \cref{sec:data}.

\paragraph{Data partitions.}
Trajectories with the same initial physical state, rounded coordinatewise to $10^{-5}$, form a source group (exact duplicate trajectories are kept in the same group). We use the initial joint positions and velocities for Cube and the full initial state for Push-T. All trajectories in a group remain in one partition. Cube has 9{,}800 groups with one trajectory each; Push-T has 185 groups with 101 trajectories each. Using seed 20261021, we randomly partition the groups in proportions of 90/5/5\% (Cube) and 70/15/15\% (Push-T); the training and selection group counts are rounded down and the remaining groups form the test partition. \Cref{tab:group_partitions} gives the resulting counts.

\begin{table}[ht]
\caption{\textbf{Additional evaluation partitions.} Counts refer to source groups and execution trajectories. Selection denotes the condition-head checkpoint-selection partition.}
\label{tab:group_partitions}
\centering\small\setlength{\tabcolsep}{4pt}
\input{tables/group_partitions.tex}
\end{table}

\paragraph{Backbones and condition heads.}
For each environment, we train HWM backbones from random initialization on the training partition only, with seeds 42, 43, and 44, using the released architecture and training schedule (eight epochs for Cube and nine for Push-T). We use the final-epoch checkpoint, specified before training, and fit normalization statistics on the training partition. Each backbone is then frozen and shared by HWM search and three condition heads initialized with seeds 0, 1, and 2. Head training uses segments from the training partition with $K=2$ (Cube) and $K=8$ (Push-T), the full three-term objective, and the architecture, optimizer, and 20{,}000-update budget in \cref{sec:implementation}; under the segment-length rule of \cref{sec:implementation}, all 8{,}820 Cube and 11{,}514 of the 13{,}029 Push-T training trajectories are eligible. Checkpoints are selected by minimum autoregressive endpoint MAE on 256 fixed windows from the selection partition (sampling seed 7), checked every 2{,}000 updates.

\paragraph{Test tasks and control.}
With sampling seed 20261022, we draw 50 execution trajectories, eligible under the same segment-length rule, uniformly without replacement from each test partition. For each trajectory, we sample a valid start uniformly on the five-step embedding grid and place the goal $10K$ environment steps later, following the head-training window rule; thus $K=2$ on Cube and $K=8$ on Push-T give 20- and 80-step start--goal gaps. These tasks cover 50 source groups on Cube and 20 on Push-T. The split and the task lists are fixed before backbone and head training; no filtering by success, geometry, or model error is applied. Compared with the benchmark assets, the Cube windows keep the 20-step gap but replace pickup-centered sampling with uniform starts, and the Push-T windows use an 80-step instead of a 74-step gap.

Each task resets the simulator to the recorded state at the start frame and uses the recorded state and observation $10K$ steps later as the goal; success is judged by the environment's released success check. Both methods use the same 50 tasks, backbone, normalization statistics, and per-task planner seeds within each backbone seed. Execution budgets are 30 steps on Cube and 90 on Push-T. Low-level planning, the schedules, and HWM search (the budgets of \main{Tab.~\ref{tab:budget}} and the one-state context) follow \cref{sec:supp_inference}, with matching weight $\beta=0$ and chunk size one.

\paragraph{Results.}
\Cref{tab:group_results} reports success rates using the aggregation in \cref{sec:supp_protocol}: three heads are averaged within each backbone, followed by the mean and SE across three backbone seeds. HWM search has one run per backbone. On the fixed task set, the SE is computed across the three backbone-level means and does not estimate variation across sampled tasks or source groups.

\begin{table}[ht]
\caption{\textbf{Success on the additional task sets (\%).} Backbones trained from scratch on the training partition (seeds 42--44; not the released checkpoints of \main{Tab.~\ref{tab:main}}). Mean $\pm$ SE across the three backbones, with three heads averaged within each backbone for ours; HWM search has one run per backbone. All runs use $\beta=0$, chunk size one, and the same 50 tasks per environment.}
\label{tab:group_results}
\centering\small\setlength{\tabcolsep}{9pt}
\input{tables/group_results.tex}
\end{table}

%% file: tables/per_head.tex
\begin{tabular}{llrrrr}
\toprule
Environment & Backbone & Head 0 & Head 1 & Head 2 & Mean \\
\midrule
Cube & 42 & 64 & 72 & 70 & 68.7 \\
Cube & 43 & 64 & 70 & 68 & 67.3 \\
Cube & 44 & 76 & 76 & 74 & 75.3 \\
\midrule
Push-T & 42 & 42 & 44 & 46 & 44.0 \\
Push-T & 43 & 50 & 52 & 60 & 54.0 \\
Push-T & 44 & 42 & 52 & 50 & 48.0 \\
\midrule
FourRoom & 42 & 100 & 100 & 100 & 100.0 \\
FourRoom & 43 & 100 & 100 & 100 & 100.0 \\
FourRoom & 44 & 100 & 100 & 100 & 100.0 \\
\midrule
AntMaze & 42 & 72 & 76 & 64 & 70.7 \\
AntMaze & 43 & 70 & 70 & 64 & 68.0 \\
AntMaze & 44 & 76 & 72 & 76 & 74.7 \\
\bottomrule
\end{tabular}

%% file: tables/per_baseline.tex
\begin{tabular}{llrrrr}
\toprule
Environment & Planner & 42 & 43 & 44 & Mean $\pm$ SE \\
\midrule
Cube & HWM search & 48 & 44 & 62 & $51.3\,{\scriptstyle\pm\,5.5}$ \\
Cube & H-JEPA search & 60 & 40 & 40 & $46.7\,{\scriptstyle\pm\,6.7}$ \\
Cube & LeWM & 24 & 26 & 28 & $26.0\,{\scriptstyle\pm\,1.2}$ \\
Cube & Linear waypoints & 4 & 2 & 8 & $4.7\,{\scriptstyle\pm\,1.8}$ \\
\midrule
Push-T & HWM search & 50 & 44 & 52 & $48.7\,{\scriptstyle\pm\,2.4}$ \\
Push-T & H-JEPA search & 32 & 52 & 38 & $40.7\,{\scriptstyle\pm\,5.9}$ \\
Push-T & LeWM & 42 & 52 & 46 & $46.7\,{\scriptstyle\pm\,2.9}$ \\
Push-T & Linear waypoints & 2 & 2 & 0 & $1.3\,{\scriptstyle\pm\,0.7}$ \\
\midrule
FourRoom & HWM search & 80 & 58 & 78 & $72.0\,{\scriptstyle\pm\,7.0}$ \\
FourRoom & H-JEPA search & 86 & 66 & 84 & $78.7\,{\scriptstyle\pm\,6.4}$ \\
FourRoom & LeWM & 42 & 44 & 34 & $40.0\,{\scriptstyle\pm\,3.1}$ \\
FourRoom & Linear waypoints & 0 & 0 & 0 & $0.0\,{\scriptstyle\pm\,0.0}$ \\
\midrule
AntMaze & HWM search & 36 & 36 & 34 & $35.3\,{\scriptstyle\pm\,0.7}$ \\
AntMaze & H-JEPA search & 40 & 34 & 38 & $37.3\,{\scriptstyle\pm\,1.8}$ \\
AntMaze & LeWM & 14 & 16 & 24 & $18.0\,{\scriptstyle\pm\,3.1}$ \\
AntMaze & Linear waypoints & 0 & 0 & 0 & $0.0\,{\scriptstyle\pm\,0.0}$ \\
\bottomrule
\end{tabular}

%% file: tables/training_cost.tex
\begin{tabular}{lrrr}
\toprule
Environment & Heads & Median (min) & Range (min) \\
\midrule
Cube & 9 & 6.9 & 6.9--8.3 \\
Push-T & 9 & 32.2 & 28.6--43.3 \\
FourRoom & 9 & 24.0 & 23.6--30.4 \\
AntMaze & 9 & 26.0 & 25.7--31.4 \\
\bottomrule
\end{tabular}

%% file: tables/ablation_cube_pusht.tex
\begin{tabular}{l@{\hskip 10pt}rrrr@{\hskip 16pt}rrrr}
\toprule
& \multicolumn{4}{c}{Cube} & \multicolumn{4}{c}{Push-T} \\
\cmidrule(lr){2-5}\cmidrule(lr){6-9}
Variant & Head 0 & Head 1 & Head 2 & Mean $\pm$ SD & Head 0 & Head 1 & Head 2 & Mean $\pm$ SD \\
\midrule
Ours (full) & 64 & 72 & 70 & $68.7\,{\scriptstyle\pm\,4.2}$ & 42 & 44 & 46 & $44.0\,{\scriptstyle\pm\,2.0}$ \\
Without teacher forcing & 68 & 66 & 68 & $67.3\,{\scriptstyle\pm\,1.2}$ & 38 & 42 & 54 & $44.7\,{\scriptstyle\pm\,8.3}$ \\
Without rollout loss & 74 & 68 & 72 & $71.3\,{\scriptstyle\pm\,3.1}$ & 36 & 36 & 28 & $33.3\,{\scriptstyle\pm\,4.6}$ \\
Without endpoint loss & 70 & 72 & 68 & $70.0\,{\scriptstyle\pm\,2.0}$ & 38 & 40 & 44 & $40.7\,{\scriptstyle\pm\,3.1}$ \\
Teacher forcing only & 68 & 70 & 74 & $70.7\,{\scriptstyle\pm\,3.1}$ & 24 & 30 & 40 & $31.3\,{\scriptstyle\pm\,8.1}$ \\
$\lambda=0.1$ & 76 & 64 & 68 & $69.3\,{\scriptstyle\pm\,6.1}$ & 42 & 40 & 46 & $42.7\,{\scriptstyle\pm\,3.1}$ \\
$\lambda=10$ & 66 & 62 & 68 & $65.3\,{\scriptstyle\pm\,3.1}$ & 20 & 26 & 22 & $22.7\,{\scriptstyle\pm\,3.1}$ \\
Short training segments & 14 & 14 & 12 & $13.3\,{\scriptstyle\pm\,1.2}$ & 20 & 24 & 32 & $25.3\,{\scriptstyle\pm\,6.1}$ \\
Long training segments & 60 & 74 & 70 & $68.0\,{\scriptstyle\pm\,7.2}$ & 26 & 16 & 16 & $19.3\,{\scriptstyle\pm\,5.8}$ \\
4D conditions via $A$ & 74 & 72 & 74 & $73.3\,{\scriptstyle\pm\,1.2}$ & 50 & 38 & 34 & $40.7\,{\scriptstyle\pm\,8.3}$ \\
Supervised lifted conditions & 68 & 70 & 68 & $68.7\,{\scriptstyle\pm\,1.2}$ & 20 & 28 & 26 & $24.7\,{\scriptstyle\pm\,4.2}$ \\
Supervised macro-actions & 74 & 70 & 72 & $72.0\,{\scriptstyle\pm\,2.0}$ & 24 & 30 & 28 & $27.3\,{\scriptstyle\pm\,3.1}$ \\
Direct subgoals & 66 & 74 & 68 & $69.3\,{\scriptstyle\pm\,4.2}$ & 46 & 50 & 42 & $46.0\,{\scriptstyle\pm\,4.0}$ \\
No head state input & 50 & 50 & 56 & $52.0\,{\scriptstyle\pm\,3.5}$ & 0 & 2 & 0 & $0.7\,{\scriptstyle\pm\,1.2}$ \\
No remaining-step input & 78 & 70 & 60 & $69.3\,{\scriptstyle\pm\,9.0}$ & 44 & 58 & 40 & $47.3\,{\scriptstyle\pm\,9.5}$ \\
\bottomrule
\end{tabular}

%% file: tables/ablation_fourroom_ant.tex
\begin{tabular}{l@{\hskip 10pt}rrrr@{\hskip 16pt}rrrr}
\toprule
& \multicolumn{4}{c}{FourRoom} & \multicolumn{4}{c}{AntMaze} \\
\cmidrule(lr){2-5}\cmidrule(lr){6-9}
Variant & Head 0 & Head 1 & Head 2 & Mean $\pm$ SD & Head 0 & Head 1 & Head 2 & Mean $\pm$ SD \\
\midrule
Ours (full) & 100 & 100 & 100 & $100.0\,{\scriptstyle\pm\,0.0}$ & 72 & 76 & 64 & $70.7\,{\scriptstyle\pm\,6.1}$ \\
Short training segments & 30 & 34 & 36 & $33.3\,{\scriptstyle\pm\,3.1}$ & -- & -- & -- & -- \\
Long training segments & 100 & 100 & 100 & $100.0\,{\scriptstyle\pm\,0.0}$ & -- & -- & -- & -- \\
4D conditions via $A$ & -- & -- & -- & -- & 76 & 66 & 70 & $70.7\,{\scriptstyle\pm\,5.0}$ \\
Direct subgoals & 98 & 98 & 100 & $98.7\,{\scriptstyle\pm\,1.2}$ & 58 & 64 & 56 & $59.3\,{\scriptstyle\pm\,4.2}$ \\
No head state input & 10 & 4 & 6 & $6.7\,{\scriptstyle\pm\,3.1}$ & 2 & 0 & 4 & $2.0\,{\scriptstyle\pm\,2.0}$ \\
\bottomrule
\end{tabular}

%% file: tables/timing_v6.tex
\begin{tabular}{llrrrrrr}
\toprule
Environment & Method & High (s) & Low (s) & Total (s) & Speedup & Low calls & High calls \\
\midrule
Cube & HWM search & 26.189 & 100.80 & 148.44 & -- & 4 & 1 \\
Cube & Ours & 0.068 & 93.65 & 115.66 & $1.28\times$ & 4 & 1 \\
\midrule
Push-T & HWM search & 505.278 & 237.39 & 757.51 & -- & 10 & 7 \\
Push-T & Ours & 0.201 & 207.31 & 221.56 & $3.42\times$ & 10 & 7 \\
\midrule
FourRoom & HWM search & 1963.936 & 768.98 & 2784.24 & -- & 22 & 19 \\
FourRoom & Ours & 0.661 & 586.35 & 634.73 & $4.39\times$ & 22 & 19 \\
\midrule
AntMaze & HWM search & 1871.047 & 991.06 & 2892.24 & -- & 36 & 31 \\
AntMaze & Ours & 0.463 & 816.15 & 846.92 & $3.41\times$ & 36 & 31 \\
\bottomrule
\end{tabular}

%% file: tables/group_partitions.tex
\begin{tabular}{lrrr}
\toprule
 & Train & Selection & Test \\
\midrule
Cube groups & 8{,}820 & 490 & 490 \\
Cube trajectories & 8{,}820 & 490 & 490 \\
\addlinespace
Push-T groups & 129 & 27 & 29 \\
Push-T trajectories & 13{,}029 & 2{,}727 & 2{,}929 \\
\bottomrule
\end{tabular}

%% file: tables/group_results.tex
\begin{tabular}{lcc}
\toprule
Environment & HWM search & Ours \\
\midrule
Cube & $88.0\,{\scriptstyle\pm\,1.2}$ & $92.7\,{\scriptstyle\pm\,1.8}$ \\
Push-T & $28.0\,{\scriptstyle\pm\,0.0}$ & $26.2\,{\scriptstyle\pm\,1.0}$ \\
\bottomrule
\end{tabular}